\pdfoutput=1

\ifdefined\REVIEW
    \documentclass[
        a4paper,
    ]{article}
    \usepackage{lineno, setspace}
    \modulolinenumbers[1]
    \pagewiselinenumbers
\else
    \documentclass[
        a4paper,
        twocolumn
    ]{article}
\fi

\usepackage{IEK10} \usepackage{natbib}

\def\IEK10{
  Forschungszentrum Jülich GmbH,
  Institute of Climate and Energy Systems,
  Energy Systems Engineering (ICE-1),
  Jülich 52425,
  Germany
}
\def\SVT{
  RWTH Aachen University,
  Process Systems Engineering (AVT.SVT),
  Aachen 52074,
  Germany
}
\def\JARA{
  JARA-CSD,
  Jülich 52425,
  Germany
}
\def\RWTH{
  RWTH Aachen University,
  Aachen 52062,
  Germany
}

\newcommand{\mytitle}{
Task-optimal data-driven surrogate models for eNMPC via differentiable simulation and optimization
}

\newcommand{\affil}{
  \begin{itemize}[leftmargin=3mm, itemsep=0mm]
    \item[$^a$]\IEK10
    \item[$^b$]\SVT
    \item[$^c$]\JARA
    \item[$^d$]\RWTH
  \end{itemize}
}

\def\firstAuthor{Daniel Mayfrank}
\newcommand{\myauthor}{
\firstAuthor$^{a,d}$\orcidlink{0009-0000-6275-0614}, 
Na Young Ahn$^{a}$\orcidlink{0009-0005-4416-2428}, 
Alexander Mitsos$^{c,a,b}$\orcidlink{0000-0003-0335-6566}, 
Manuel Dahmen$^{a,*}$\orcidlink{0000-0003-2757-5253} }
\newcommand{\correspondingauthor}{Manuel Dahmen, \IEK10 \newline E-mail: m.dahmen@fz-juelich.de}
\author{\myauthor}

\usepackage[
  colorlinks,
  linkcolor=blue,
  citecolor=blue,
  urlcolor=blue,
  pdftitle={\mytitle},
  pdfauthor={\firstAuthor}
]{hyperref}
\usepackage[capitalise, nameinlink]{cleveref}
\crefname{table}{Tab.}{Tab.}

\usepackage{orcidlink}

\newcommand{\setpgfexternalcounter}[1]{
  \makeatletter \pgfkeysgetvalue{/tikz/external/figure name}\myexternalname
  \expandafter\gdef\csname c@tikzext@no@\myexternalname\endcsname{#1}\makeatother
}

\begin{document}

\ifx\REVIEW\undefined
\twocolumn[
\begin{@twocolumnfalse}
\fi
  \thispagestyle{firststyle}

  \begin{center}
    \begin{large}
      \textbf{\mytitle}
    \end{large} \\
    \myauthor
  \end{center}

  \vspace{0.5cm}

  \begin{footnotesize}
    \affil
  \end{footnotesize}

  \vspace{0.5cm}

    Mechanistic dynamic process models may be too computationally expensive to be usable as part of a real-time capable predictive controller. We present a method for end-to-end learning of Koopman \textit{surrogate} models for optimal performance in a specific control task. In contrast to previous contributions that employ standard reinforcement learning (RL) algorithms, we use a training algorithm that exploits the differentiability of environments based on mechanistic simulation models to aid the policy optimization. We evaluate the performance of our method by comparing it to that of other training algorithms on an existing economic nonlinear model predictive control (eNMPC) case study of a continuous stirred-tank reactor (CSTR) model. Compared to the benchmark methods, our method produces similar economic performance while eliminating constraint violations. Thus, for this case study, our method outperforms the others and offers a promising path toward more performant controllers that employ dynamic surrogate models.

  \vspace*{5mm}
\ifx\REVIEW\undefined
\end{@twocolumnfalse}
]
\fi

\newpage

\section{Introduction}\label{sec:intro}

    Economic model predictive control (eNMPC) is a control strategy that uses a dynamic process model to predict the system behavior and make real-time control decisions by repeatedly solving an optimal control problem (OCP) in a rolling horizon fashion. Whereas traditional model predictive control (MPC) focuses on following reference trajectories, eNMPC aims to directly optimize economic process performance by integrating an economic objective into the OCP. eNMPC relies on a sufficiently accurate dynamic model of the process. Unfortunately, for high-dimensional nonlinear systems, the computational burden of solving the resulting OCPs can render eNMPC computationally intractable. In such cases, data-driven \textit{surrogate} models for computationally expensive mechanistic dynamic models can enable real-time \mbox{eNMPC} by reducing the computational burden of solving the underlying OCPs (\cite{mcbride2019overview}).

\begin{figure}[ht]
        \centering
        \subfloat[a][]{\includegraphics[width=0.37\paperwidth]{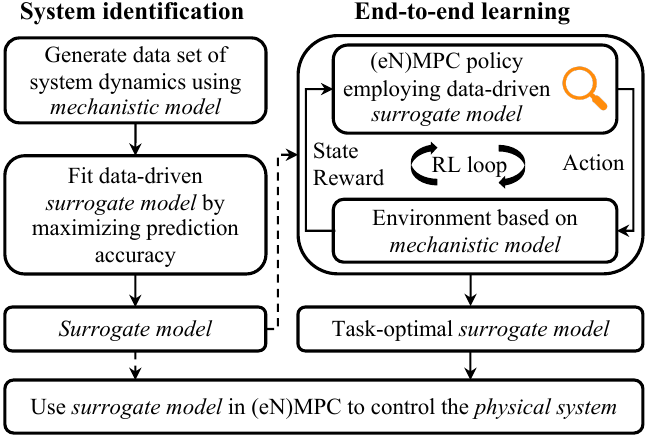} \label{fig:SI_vs_E2E}} \\
        \subfloat[b][]{\includegraphics[width=0.28\paperwidth]{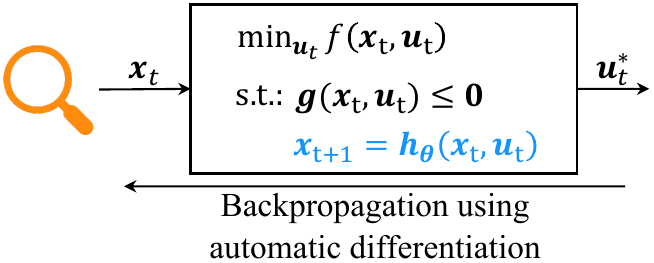} \label{fig:differentiable_eNMPC}}
        \caption{(a) Comparison of two paradigms for the training of data-driven dynamic surrogate models for use in eNMPC. (b) The differentiable eNMPC policy takes as input the current state $\bm{x}_t$ and computes the optimal control action $\bm{u}^{*}_{t}$ based on a cost function $f$, inequality constraints $\bm{g}$, and the learnable discrete-time dynamic surrogate model $\bm{h_\theta}$ (highlighted in blue font), which is parameterized by $\bm{\theta}$.} \label{fig:introduction}
    \end{figure}

    Recent articles (e.g., \cite{gros2019data, mayfrank2024end}) have established end-to-end reinforcement learning (RL) of dynamic surrogate models as way to train models for optimal performance in a specific (control) task (see Fig. \ref{fig:SI_vs_E2E}).
    Task-optimal dynamic models for control can be learned by viewing a dynamic model and its learnable parameters as part of a differentiable policy (see Fig. \ref{fig:differentiable_eNMPC}). This policy consists of the dynamic model and a differentiable optimal control algorithm. Various methods for turning (e)NMPC controllers into differentiable and thereby learnable policies have been developed, see, e.g., (\cite{amos2018differentiable, gros2019data, mayfrank2024end}). These methods do not depend on any specific policy optimization algorithm. End-to-end RL of surrogate models may yield increased performance of the resulting eNMPCs regarding the respective control objective, e.g., the minimization of operating costs while avoiding constraint violations (\cite{mayfrank2024end}).

    RL algorithms are a class of policy optimization algorithms that enable learning of optimal controllers or dynamic models for use in eNMPC through trial-and-error actuation of real-world or simulated environments. Standard RL algorithms view environments as black boxes and do not use derivative information regarding the environment dynamics or the reward signals, even though many RL publications use simulated environments where analytical gradients of dynamics and rewards could be available. Policy gradient algorithms are the most commonly used class of RL algorithms for end-to-end learning of dynamic models for control, see, e.g., (\cite{gros2019data}). However, fundamental issues arise from the fact that policy gradient algorithms do \textit{not} leverage analytical gradients from the environment. These issues concern both a lacking understanding of the algorithms' empirical behavior (\cite{ilyas2018closer, wu2022understanding}) and their performance (\cite{islam2017reproducibility, henderson2018did}).

    Recently, however, policy optimization algorithms that leverage the derivative information from simulated environments were designed, e.g., the Short-Horizon Actor-Critic (SHAC) algorithm (\cite{xu2022accelerated}). These algorithms manage to avoid the well-known problems of Backpropagation Through Time (BPTT) (\cite{werbos1990backpropagation}), i.e., noisy optimization landscapes and exploding/vanishing gradients (\cite{xu2022accelerated}), and have shown enhanced training wall-clock time efficiency and increased terminal performance compared to state-of-the-art RL algorithms that do not exploit derivative information from the environment. These algorithmic advances could also benefit the learning of dynamic surrogate models for (eN)MPC if the mechanistic simulation model is differentiable. To this end, differentiable simulators, e.g., (\cite{chen2018neuralode}), can be used to construct simulated RL environments with automatically differentiable dynamics and reward functions, thus enabling the use of analytic gradients for policy optimization. Nevertheless, policy optimization using differentiable environments has yet to be established for the learning of dynamic surrogate models for (eN)MPC.

    By combining our previously proposed method for end-to-end learning of task-optimal Koopman models in (e)NMPC applications (\cite{mayfrank2024end}) with the SHAC algorithm (\cite{xu2022accelerated}), we present a method for end-to-end optimization of Koopman surrogate models. Crucially, our method exploits the differentiability of simulated environments, distinguishing it from previous contributions, which are based on RL (\cite{gros2019data, mayfrank2024end}) or imitation learning (\cite{amos2018differentiable}). We evaluate the resulting control performance on an eNMPC case study derived from a literature-known continuous stirred-tank reactor model (\cite{flores2006simultaneous}). We compare the performance to that of eNMPCs employing Koopman surrogate models that were trained either using (\textit{i}) system identification or (\textit{ii}) RL. We find that the novel combination of a Koopman-eNMPC trained using SHAC exhibits superior performance compared to the other options. This finding confirms our expectation that the advantages of policy optimization algorithms that leverage derivative information from differentiable environments can apply to the end-to-end training of dynamic surrogate models for predictive control applications. Thus, our work constitutes a step towards more performant real-time capable optimal control policies for large-scale, nonlinear systems, where optimal control policies based on a mechanistic model are not real-time capable.

    We structure the remainder of this paper as follows: Section \ref{sec:method} presents our method. Section \ref{sec:num_experiments} showcases the performance of our method on a simulated case study and discusses the results. Section \ref{sec:conclusion} draws some final conclusions.

\section{Method}\label{sec:method}
    From a methodological perspective, the core contribution of this work is combining: (i) the SHAC algorithm (\cite{xu2022accelerated}), a policy optimization algorithm that leverages derivative information from a differentiable simulation environment, (ii) an extension of Koopman theory to controlled systems by Korda and Mezić (\cite{korda2018linear}) that results in convex OCPs, and (iii) our previously published (\cite{mayfrank2024end}) method for turning Koopman-(e)NMPCs into differentiable policies. We refer the reader to the aforementioned publications for a detailed description of the theoretical background of this work.

    We adopt an RL perspective (\cite{sutton2018reinforcement}) on policy optimization problems. Herein, the control problem is represented by a discrete-time Markov Decision Process (MDP) with associated states $\bm{x}_t \in \mathbb{R}^n$ and control inputs $\bm{u}_t \in \mathbb{R}^m$, a transition function $\bm{\mathcal{F}}: \mathbb R^n \times \mathbb R^m \to \mathbb R^n$, ${\bm{x}_{t+1} = \bm{\mathcal{F}}(\bm{x}_{t},\bm{u}_{t})}$, and a scalar reward function $\mathcal{R}: \mathbb R^n \times \mathbb R^m \to \mathbb{R}$, ${r_{t+1} = \mathcal{R}(\bm{x}_{t+1},\bm{u}_{t})}$. An \textit{environment} is the MDP of a specific RL problem. An \textit{episode} refers to a sequence of interactions between a policy and its environment, starting from an initial state, involving a series of control inputs, and leading to a terminal state. A policy $\bm{\pi}_{\bm{\theta}}(\bm{u}_{t}|\bm{x}_{t})\colon \mathbb{R}^n \to \mathbb{R}^m$ is a function, parameterized by $\bm{\theta}$, mapping states $\bm{x}_{t}$ to (probability distributions over) control inputs $\bm{u}_{t}$. The goal of policy optimization is to maximize the expected future sum of rewards.

    \begin{figure}[htb]
    	\centering
    	\includegraphics[width=0.37\paperwidth]{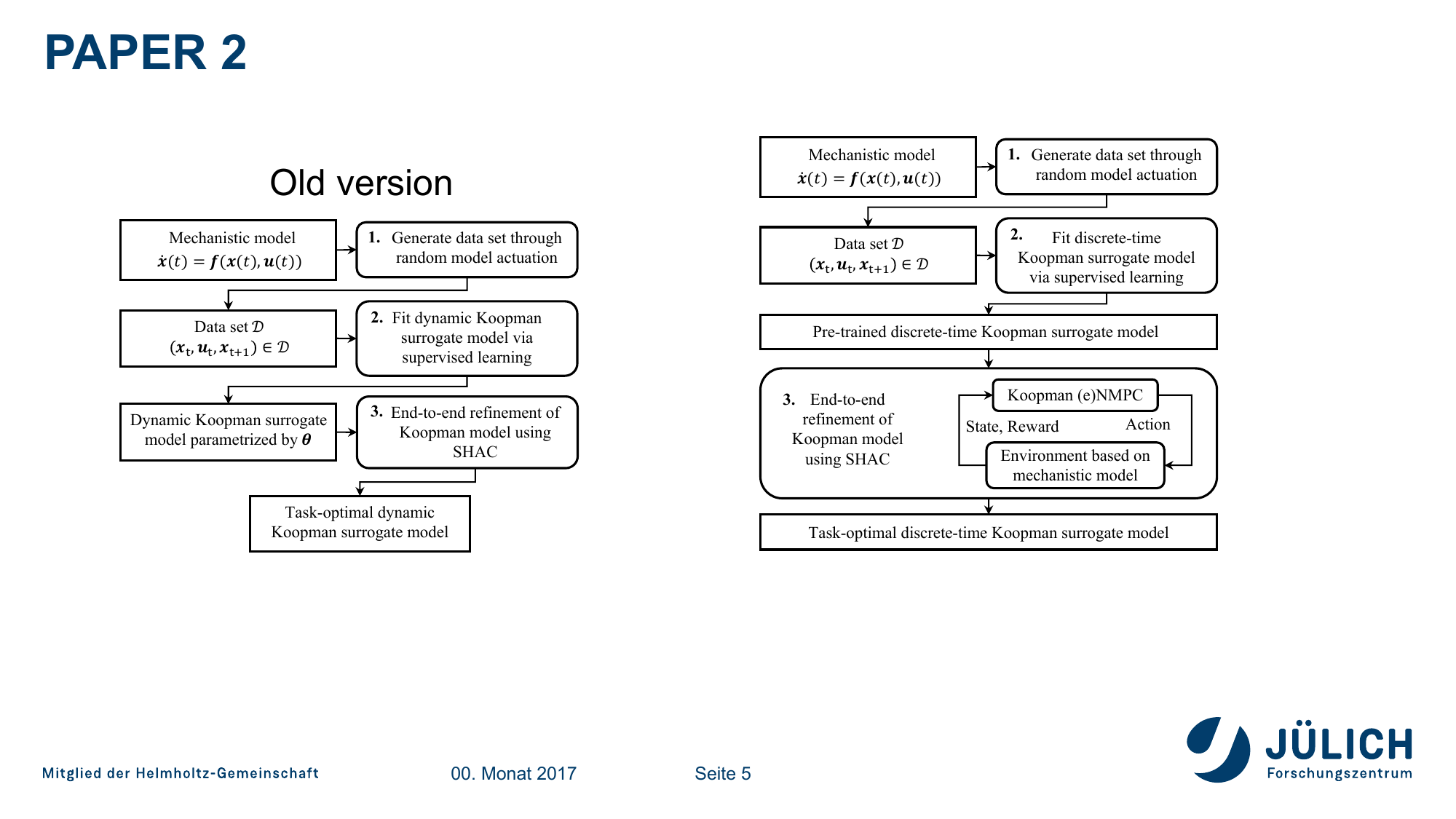}
    	\caption{Workflow from mechanistic model to task-optimal dynamic Koopman surrogate model. Adapted from \cite{mayfrank2024end}.}
    	\label{fig:workflow}
    \end{figure}
    \begin{figure*}[ht]
    	\centering
    	\includegraphics[width=0.75\paperwidth]{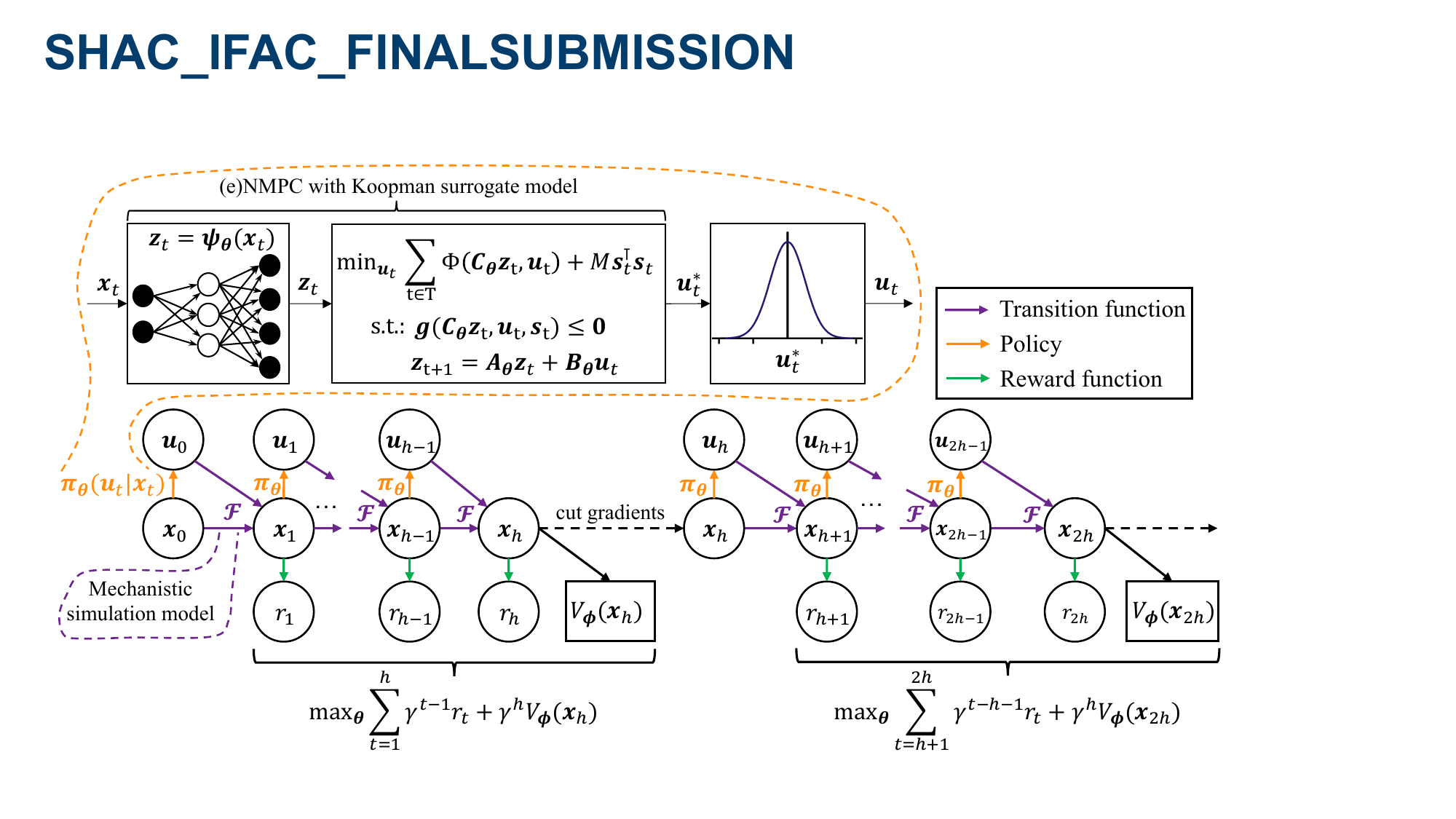}
    	\caption{Using SHAC to train a task-optimal Koopman surrogate model for the transition function $\bm{\mathcal{F}}$. This figure can be interpreted as a SHAC-specific unrolled version of the typical RL loop shown in the third step in Fig. \ref{fig:workflow}. The policy is optimized by adjusting the parameters $\bm{\theta}$ of the dynamic Koopman surrogate model. $\Phi$ is a convex function for the stage cost of the objective function. To ensure the feasibility of the resulting optimal control problems, we add slack variables $\bm{s}_t$ to the state bounds (\cite{mayfrank2024end}). Their use is penalized quadratically using a penalty factor $M$. Due to the use of PyTorch and \textit{cvxpylayers} (\cite{Agrawal2019differentiable}), the output $\bm{u}_t$ of the policy is differentiable with respect to $\bm{x}_t$ and $\bm{\theta}$. The critic is a feedforward neural network with trainable parameters $\bm{\phi}$. To increase the clarity of the figure, we omit the direct dependence of $r_{t+1}$ with respect to $\bm{u}_t$.}
    	\label{fig:SHAC}
    \end{figure*}
    We aim to exploit the differentiability of continuous-time mechanistic models that can be represented as ordinary differential equation (ODE) systems, i.e., 
    \begin{equation}
        \dot{\bm{x}}(t) = \bm{f}(\bm{x}(t), \bm{u}(t)) ,\label{eq:mechanistic_model}
    \end{equation}
    for the end-to-end learning of task-optimal discrete-time dynamic surrogate models. In our previous publication (\cite{mayfrank2024end}), we present a method for end-to-end RL of data-driven Koopman models for optimal performance in (e)NMPC applications, based on viewing the predictive controller as a differentiable policy and training it using RL. This method is independent of the specific policy optimization algorithm. Therefore, by replacing the RL algorithm (\cite{schulman2017proximal}) with SHAC (\cite{xu2022accelerated}), we can take advantage of policy optimization algorithms that exploit the differentiability of simulated environments in the learning of surrogate models for dynamic optimization. Analogous to the approach we take in (\cite{mayfrank2024end}), the overall workflow (visualized in Fig. \ref{fig:workflow}) from a mechanistic model to a task-optimal dynamic Koopman surrogate model consists of three steps: (\textit{i}) We generate a data set of the system dynamics by simulating the mechanistic model using randomly generated control inputs. (\textit{ii}) Following the model structure proposed by \cite{korda2018linear}, we fit a Koopman model (\cite{koopman1931hamiltonian}) with learnable parameters $\bm{\theta}$ to the data. (\textit{iii}) Using the mechanistic process model (Eq. \ref{eq:mechanistic_model}) and a differentiable simulator (\cite{chen2018neuralode}), we construct a differentiable RL environment whose reward formulation incentivizes task-optimal controller performance on a specific control task. For instance, in an eNMPC application, the task-specific reward should depend on operating costs and potential constraint violations, not on the prediction accuracy of the dynamic model, which is used as part of the predictive controller. Using the differentiable environment, we fine-tune the Koopman model for task-optimal performance as part of a predictive controller. Fig. \ref{fig:SHAC} provides a conceptual sketch of this fine-tuning process. To ensure exploration in the training process, we add Gaussian noise to the -- otherwise deterministic -- output of the Koopman eNMPC policy. For a detailed description of steps one and two in Fig. \ref{fig:workflow} (data generation and SI) and how to construct a differentiable eNMPC policy from a Koopman surrogate model, we refer the reader to our previous work (\cite{mayfrank2024end}).

\section{Numerical experiments}\label{sec:num_experiments}

\subsection{Case study description}\label{sec:case_study}
    Following our previous work (\cite{mayfrank2024end}), we consider a dimensionless benchmark continuous stirred-tank reactor (CSTR) model (\cite{flores2006simultaneous}) that consists of two states (product concentration $c$ and temperature $T$), two control inputs (production rate $\rho$ and coolant flow rate $F$), and two nonlinear ordinary differential equations:
    \begin{subequations} 
    	\begin{align*}
            \dot{c}(t) = &\:(1-c(t)) \dfrac{\rho(t)}{V} - c(t)ke^{-\frac{N}{T(t)}},\\
            \begin{split}
                \dot{T}(t) = &\:(T_f - T(t))\dfrac{\rho(t)}{V} + c(t)ke^{-\frac{N}{T(t)}}\\
                &\:- F(t) \alpha_c (T(t)-T_c)
            \end{split}
    	\end{align*}
    \end{subequations}
    The model has a steady state at $c_{\text{ss}}=0.1367$, $T_{\text{ss}}=0.7293$, $\rho_{\text{ss}}=1.0\frac{1}{\text{h}}$, $F_{\text{ss}}=390.0\frac{1}{\text{h}}$. Based on the model, we construct an eNMPC application by assuming that the electric power consumption is proportional to the coolant flow rate $F$, enabling production cost savings by shifting process cooling to intervals with comparatively low electricity prices. Given price predictions, the goal is to minimize the operating costs while adhering to process constraints. The state variables and the control inputs are subject to box constraints ($0.9 c_{\text{ss}} \leq c \leq 1.1 c_{\text{ss}}$, $0.8 T_{\text{ss}} \leq T \leq 1.2 T_{\text{ss}}$, $0.8\frac{1}{\text{h}} \leq \rho \leq 1.2 \frac{1}{\text{h}}$, and $0.0\frac{1}{\text{h}} \leq F \leq 700.0 \frac{1}{\text{h}}$). We include a product storage with a maximum capacity of six hours of steady-state production to enable flexible production. To match the hourly structure of the day-ahead electricity market, we choose control steps of length $\Delta t_{\text{ctrl}} = 60$ minutes. A more detailed case study description, including the model parameters, is given in \cite{mayfrank2024end}.

\subsection{Training setup}\label{sec:training_setup}
    We compare the performance of the following three training paradigms:
    \begin{enumerate}
        \item \textit{Koopman-SI}: eNMPC controller using a Koopman surrogate model trained solely using SI.
        \item \textit{Koopman-PPO}: eNMPC controller using a Koopman surrogate model pretrained using SI and refined for task-optimal performance using the state-of-the-art policy gradient algorithm Proximal Policy Optimization (PPO) (\cite{schulman2017proximal}), like we did in \cite{mayfrank2024end}.
        \item \textit{Koopman-SHAC} (main contribution of this work): eNMPC controller using a Koopman surrogate model pretrained using SI and refined for task-optimal performance using the SHAC algorithm (\cite{xu2022accelerated}).
    \end{enumerate}

    Our goal is to train dynamic surrogate models of fixed size for optimal performance as part of eNMPC. All results presented in Subsection \ref{sec:results} are obtained using a Koopman model with a latent space dimensionality of eight, i.e., $\bm{A}_{\bm{\theta}} \in \mathbb{R}^{8 \times 8}, \bm{B}_{\bm{\theta}} \in \mathbb{R}^{8 \times 2}, \bm{C}_{\bm{\theta}} \in \mathbb{R}^{2 \times 8}$, and an encoder $\bm{\psi}\colon \mathbb{R}^2 \to \mathbb{R}^8$ in the form of a feedforward neural network with two hidden layers, four and six neurons, respectively and hyperbolic tangent activation functions. We use an eNMPC horizon of nine hours.

    We use the same data set as in (\cite{mayfrank2024end}) for the SI pretraining of the Koopman surrogate model. This data set consists of 84 trajectories, each of a length of 5 days, i.e., 480 time steps, using a step length of 15 minutes. Of those 84 trajectories, we use 63 for training and the remaining 21 for validation. Then, we perform SI of the Koopman model in a similar way as described in (\cite{mayfrank2024end}). We repeat SI ten times using random seeds. We use the model with the lowest validation loss for the \textit{Koopman-SI} controller. The same model is used as the initial guess when training the \textit{Koopman-PPO} and \textit{Koopman-SHAC} controllers.

    In order to use a policy optimization algorithm such as SHAC, which makes use of derivative information from a simulated environment, not only the dynamic model but also the reward function must be differentiable. For our case study, we choose a reward function that calculates the reward at time step $t$ based on whether any constraints were violated at that time step, and on the electricity cost savings compared to the steady-state production at nominal rate between $t-1$ and $t$. The constraint component $r^{\text{con}}_{t}$ of the reward quadratically penalizes violations of the bounds of $c$, $T$, and the product storage, i.e., $r^{\text{con}}_{t} \geq 0$, and $r^{\text{con}}_{t} = 0$ if no constraint violation occurs at $t$. The cost-component $r^{\text{cost}}_{t}$ of the reward is calculated by taking the difference between the cost at nominal production and the actual production cost between $t-1$ and $t$, i.e.,
    \begin{equation*}
        r^{\text{cost}}_{t} = (F_{\text{ss}} - F_{t-1}) \cdot p_{t-1} \cdot \Delta t_{\text{ctrl}},
    \end{equation*}
    where $p_{t-1}$ is the electricity price between $t-1$ and $t$, and $\Delta t_{\text{ctrl}}$ is the time between $t-1$ and $t$ for which the controls are held constant. We balance the influence of the two components on the overall reward using a hyperparameter $\alpha$:
    \begin{equation*}
        r_{t} = \alpha \cdot r^{\text{cost}}_{t} - r^{\text{con}}_{t}
    \end{equation*}

    We train the policies using day-ahead electricity prices from the Austrian market from March 29, 2015, to March 25, 2018 (\cite{open_power_system_data_2020}). Using random seeds, we repeat the controller training ten times for each combination of policy and training algorithm (except for the \textit{Koopman-SI} controller, whose Koopman model is not trained any further after SI).

    We use the same hyperparameters for \textit{Koopman-PPO} and \textit{Koopman-SHAC} wherever possible, i.e., for all hyperparameters which are not part of PPO or SHAC. For the hyperparameters of PPO and SHAC, we do not perform extensive hyperparameter tuning. Instead, we mostly rely on the standard values. In both algorithms, we use the Adam optimizer with a small learning rate of $10^{-5}$ as the behavior of the Koopman eNMPCs is highly sensitive to small changes in the parameters. We used the \textit{Stable-Baselines3} (\cite{stable-baselines3}) implementation of PPO but implemented our own version of SHAC following the description in (\cite{xu2022accelerated}). All code used for training the controllers, including the hyperparameters that were used to obtain the results presented in Subsection \ref{sec:results}, is publicly available\footnote{\url{https://jugit.fz-juelich.de/iek-10/public/optimization/shac4koopmanenmpc}}. In addition to the code used to obtain the results presented in the following subsection, the linked repository also contains code for training pure neural network policies for our case study using PPO and SHAC. As the results of the neural network policies do not influence the narrative of this work in a meaningful way we do not discuss these results.

\subsection{Results}\label{sec:results}
    For \textit{Koopman-PPO} and \textit{Koopman-SHAC}, we identify the controller (and the associated set of parameters) that achieved the highest average reward between two consecutive parameter updates. We test their performance and that of the \textit{Koopman-SI} controller on a continuous roughly half-year-long test episode using Austrian day ahead electricity price data from March 26, 2018, to September 30, 2018 (\cite{open_power_system_data_2020}). Unlike the training process, we perform this testing without exploration, i.e., we waive adding Gaussian noise to the controller output (cf. Fig. \ref{fig:SHAC}). We initialize the test episode for each controller at the steady state of the CSTR and with empty product storage. The results are presented in Table \ref{tab:test_results}. The trajectories produced by all controllers exhibit an intuitive inverse relationship between electricity prices and coolant flow rate.

    \begin{table}[ht]
        \centering
        \setlength{\extrarowheight}{0.05cm}
        \caption{Test results: The economic cost is stated relative to the nominal production cost, i.e., we report the cost incurred by the respective controller, divided by the cost incurred by steady-state production at nominal rate given the same electricity price trajectory. The percentage of control steps that result in constraint violations is given. The average size of a constraint violation is given relative to the size of the feasible range of the variable whose bound was violated.}
        \label{tab:test_results}
        \resizebox{0.37\paperwidth}{!}{
        \begin{tabular}{l|rrr}
            \toprule
                                  & \makecell[r]{Economic\\cost} & \makecell[r]{Constr.\\viols. [\%]} & \makecell[r]{Avg. size\\constr. viol.}  \\ \hline
            \textit{Koopman-SI}   &$\bm{0.88}$&        $19.88$ &           $5.1 \cdot 10^{-2}$ \\
            \textit{Koopman-PPO}  &$0.90$&             $17.57$ &           $1.3 \cdot 10^{-2}$ \\
            \textit{Koopman-SHAC} &$0.90$&          $\bm{0.0}$ &           $\bm{-} \qquad $ \\
            \bottomrule
        \end{tabular}
        }
    \end{table}
    As can be seen from Table \ref{tab:test_results}, \textit{Koopman-SHAC} is the only controller that does not produce any constraint violations. \textit{Koopman-SI} achieves slightly higher cost savings than \textit{Koopman-SHAC}, however, it frequently produces constraint violations. Therefore, we consider \textit{Koopman-SHAC} preferable in most real-world applications where constraint-satisfaction is of high importance.

    \begin{figure}[ht]
    	\centering
    	\includegraphics[width=0.38\paperwidth]{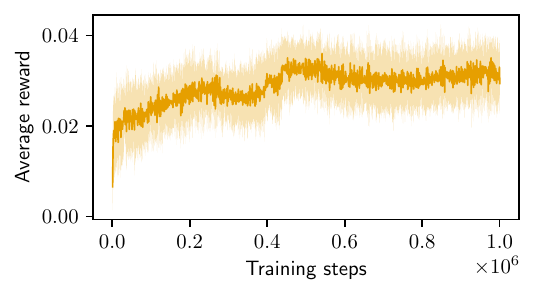}
    	\caption{Learning progress in the \textit{Koopman-SHAC} training runs. The dark orange line indicates the running average reward over the previous 1024 steps in the environment, averaged over all ten training runs. The light orange region indicates one standard deviation of the performance variance between the training runs.}
    	\label{fig:SHAC_learning_curve}
    \end{figure}

    It is noticeable that \textit{Koopman-SI} and \textit{Koopman-PPO} cause substantially more constraint violations than the \textit{Koopman-SHAC} controller. In the case of \textit{Koopman-SI}, this is not surprising since the employed Koopman surrogate model received no end-to-end training for task-optimal performance. Here, frequent but minor constraint violations show that the controller is trying to operate at the borders of the feasible range, which is normal behavior for a predictive controller. However, in the case of \textit{Koopman-PPO} the results are unsatisfactory and rather unexpected. Here, the end-to-end training reduced the average size of the constraint violations by roughly a factor of four, but only led to a small improvement in the frequency of constraint violations. We observe that some training runs did not improve performance compared to \textit{Koopman-SI} at all. Moreover, those training runs that did improve performance did not show stable convergence to high rewards. The unstable convergence of the \textit{Koopman-PPO} controllers is in line with the results in our previous work (\cite{mayfrank2024end}). There, we used a non-differentiable reward formulation with a high constant penalty incurred by any constraint violation. Using that reward formulation, we managed to substantially reduce the frequency of constraint violations unlike here. However, that approach (\cite{mayfrank2024end}) severely affected the resulting economic performance and thus produced overly conservative eNMPC controllers. In contrast to \textit{Koopman-PPO}, \textit{Koopman-SHAC} exhibits a relatively stable convergence to high rewards in our case study (Fig. \ref{fig:SHAC_learning_curve}) and produces superior terminal performance (Tab. \ref{tab:test_results}). The control performance improves relatively evenly in all ten training runs without ever dropping for extended periods.

    Due to the small size of the case study under consideration, a detailed analysis of the training runtimes would provide little insight about the expected runtime on a control problem of more practical relevance. Therefore, we leave such an analysis for future work on larger systems.

\subsubsection{Policy gradient analysis}\label{sec:policy_grad_analysis}
    The fundamental difference between the two policy optimization algorithms SHAC (\cite{xu2022accelerated}) and PPO (\cite{schulman2017proximal}) is that SHAC utilizes analytic policy gradients from an automatically differentiable environment, whereas PPO estimates policy gradients via the policy gradient theorem (\cite{sutton2018reinforcement}). \cite{ilyas2018closer} show that given common and practical hyperparameter configurations, the PPO-estimated policy gradients can incur a high variance which can lead to unstable training convergence, as we observe in our case. We follow the approach by \cite{ilyas2018closer} to analyze empirically whether there is a meaningful difference in the variance of the policy gradients produced by PPO and SHAC when applied to our case study. To this end, we investigate how similar the direction of multiple policy gradients are given a fixed policy parameterization. We fix the policy parameters to that of the \textit{Koopman-SI} controller for this analysis. Then, for PPO and SHAC, we fit the critic to this policy \textit{without} updating the policy. Finally, for both algorithms and still without updating the policy, we record the policy gradients of 100 optimization steps. We calculate how similar the 100 recorded gradients are by computing their average pairwise cosine similarity. The cosine similarity is a measure of the similarity of two vectors which only depends on their direction, not on their length. It takes a value of one if both vectors point in the same direction, zero if they are orthogonal to each other, and minus one if they point in exactly opposite directions. PPO produces an average cosine similarity of $0.22$, whereas SHAC results in an average cosine similarity of $0.94$. Thus, the variance in the direction of policy gradients is much higher for PPO than for SHAC, which might explain the instable convergence to high rewards and thus the relatively bad performance of \textit{Koopman-PPO} (see Tab. \ref{tab:test_results}).

\section{Conclusion}\label{sec:conclusion}
    We combine our previously published method for turning Koopman-(e)NMPC controllers into differentiable policies (\cite{mayfrank2024end}) with the policy optimization algorithm SHAC (\cite{xu2022accelerated}). Our approach leverages derivative information from automatically differentiable simulated environments, differentiating it from previously published methods for end-to-end training of dynamic models for control. We find that SHAC produces a stable convergence to high control performance across all independent training instances, translating to superior control performance compared to our previously published approach (\cite{mayfrank2024end}) which was based on the PPO algorithm. Note that even though our method achieves perfect constraint satisfaction in our case study, full adherence to constraints can in general not be expected.

    The results can be understood as a successful proof of concept. By training data-driven surrogate models for optimal performance in a specific control task, our method utilizes the representational capacity of the model efficiently, thus avoiding unnecessarily large and computationally expensive surrogate models. We view our approach as a promising avenue toward more performant real-time capable data-driven (e)NMPCs. Still, the computational burden of backpropagation through mechanistic simulations and optimal control problems and thus the cost of each training iteration is naturally linked to the size of the mechanistic simulation model, meaning that our method could become computationally intractable for very large models. Thus, future work should investigate the application of our method to larger mechanistic simulation models and more challenging control problems, presumably necessitating training for more iterations.

\section*{Declaration of Competing Interest}
We have no conflict of interest.

\section*{Acknowledgements}
This work was performed as part of the Helmholtz School for Data Science in Life, Earth and Energy (HDS-LEE) and received funding from the Helmholtz Association of German Research Centers.

We thank Jan C. Schulze (Process Systems Engineering (AVT.SVT), RWTH Aachen University, 52074 Aachen, Germany) for fruitful discussions and valuable feedback.

\appendix

  

\end{document}